\documentclass[sigconf]{acmart}

\usepackage{booktabs} 

\setcopyright{acmlicensed}

\acmConference[HRI2018]{Workshop on Personal Robots for Exercising and Coaching at the HRI 2018}{March 2018}{Chicago, Illinois, 2018} 

\editor{Sebastian Schneider}
\editor{Sascha Griffiths}
\editor{Carlos A. Cifuentes G.}
\editor{Stefan Wermter}
\editor{Britta Wrede}

\begin{document}
\title{Exercise with Social Robots: Companion or Coach?}

\author{Sascha Griffiths, Tayfun Alpay, Alexander Sutherland, Matthias Kerzel, Manfred Eppe, Erik Strahl, Stefan Wermter}
\affiliation{%
  \institution{Knowledge Technology\\ Universit\"{a}t Hamburg}
  \streetaddress{Vogt Koelln Str. 30}
  \city{Hamburg} 
  \country{Germany}
}
\email{griffiths, alpay, sutherland, kerzel, eppe, strahl, wermter@informatik.uni-hamburg.de}





\renewcommand{\shortauthors}{S. Griffiths et al.}


\begin{abstract}
In this paper, we investigate the roles that social robots can take in physical exercise with human partners. In related work, robots or virtual intelligent agents take the role of a coach or instructor whereas in other approaches they are used as motivational aids. These are two ``paradigms'', so to speak, within the small but growing area of robots for social exercise. We designed an online questionnaire to test whether the preferred role in which people want to see robots would be the companion or the coach. The questionnaire asks people to imagine working out with a robot with the help of three utilized questionnaires: (1) CART-Q which is used for judging coach-athlete relationships, (2) the mind perception questionnaire and (3) the System Usability Scale (SUS). We present the methodology, some preliminary results as well as our intended future work on personal robots for coaching. 
\end{abstract}

%
%



\maketitle

\section{Introduction}
There is a growing amount of people who are in need of movement due to living a largely sedentary lifestyle. Figures on this indicate that this leads to problems both at a young age and in later stages in life \cite{WHOPhysi81:online}. Leading consequences are heart disease and back problems. 

Modern technology has been adopted as an approach toward tackling this problem. The opportunities offered by robots can contribute to confronting this social need. Alternative approaches also include technological aid offered by fitness videos, nowadays but also newer technology such as video games \cite{biddiss2010active} and intelligent virtual agents (IVA)~\cite{cassell2000embodied}. Exercise videos have been around since the early 1980s and were largely popularized by a pioneering video of the genre which featured the actress Jane Fonda \cite{fleming2004effects}; this media-assisted form of a physical exercise tool can, therefore, be regarded as the oldest type of exercise aid in the list above. One obvious drawback is the limited amount of content that such videos provide. Television and video, in comparison to modern multimedia enhanced with artificial intelligence, can also be regarded as ``passive companions'' whereas video games, IVAs and social robots can be considered ``active companions''. The latter offer interaction in addition to action in this application field\footnote{The contrast between television as a ``passive companion'' as opposed to social robots as ``active companions'' was discussed by Dr Tomotaka Takahashi in his 2011 keynote at the International Conference of Social Robotics (ICSR). \url{http://icsoro.org/icsr2011/}}. It is for these two reasons, \textit{extendable content} and \textit{interactivity}, which make robots an interesting option for future solutions to the perils of sedentary lifestyles. It is the latter component that we will focus on in the following.

There is a small but growing body of literature on social and personal robots used for physical exercise and rehabilitation (see \ref{sec:robotCoach}). The promise is that robots can serve as tools for exercise in contexts where the social environment in form of another person or a group cannot be provided. Having another person present can have positive effects such as more positive affective states during exercise \cite{dunton2015momentary}, fewer injuries, reduced stress levels \cite{plante2001does}, increased performance \cite{feltz2011buddy}, or simply meeting one's goals \cite{rackow2015received}. Thus, in lieu of a human partner, technology could help to counter the adverse effects of too little exercise or incorrect exercise on one's own.

Prior research on dyads of human coaches and athletes has shown that a positive perception of a relationship is correlated with the pair's own individual personality traits \cite{jackson2011personality}. Significant differences between extroversion and openness lead to a lower, i.e. less trusting and less close, perception of the dyadic relationship than coach-athlete pairings possessing a high concordance of these traits. Pairings that had a partner high in agreeableness and conscientiousness reported having a more favorable relationship with their counterpart. 

These findings can function as markers for designing HRI strategies within the context of coaching and having robots in an advisory position. Using these markers, it is possible to assume that robots should be designed with high agreeableness and high conscientiousness in mind, while adopting a more personalized approach when considering how to act, by attempting to match their subjects extroversion and openness levels. However, given that robots are viewed as having less agency, there may exist some discrepancies between how the ideal robot coach is expected to act and how an ideal human coach may be expected to act.

The question is what type of role artificial partners, particularly social robots, should take in the training regiment of future users. There are three main lines to be discerned so far: (1) monitoring the exercise \cite{parisi2016human}, (2) delivering coaching during exercise \cite{Nguyen2016R2IISHRIC}, or (3) motivating during exercise \cite{schneider2016motivational}, with the latter being a component of the concept of having full-blown robot fitness companions \cite{sussenbach2014robot}. In terms of ``active companions'' and making full use of the embodiment offered by robot companions \cite{dautenhahn2005robot}, we see especially (2) and (3) as particularly promising for the future of technological aids for exercise. 

In the following, we will ask the question of whether the preferred role of a social robot in exercise contexts should be that of a fitness companion or a fitness coach. For this purpose, we conducted a study using an online questionnaire. We discuss the related work, both in relation to the premise of using robots and IVAs in exercise scenarios and our methodology in the following section. After that, we explain our methods with a particular focus on the structure of the questionnaire. Our results will be presented in the next section, followed by the discussion of these. Finally, we will conclude with a statement about social robots as fitness companions versus fitness coaches. 

\section{Related Work}
\subsection{Robots and IVAs in Physical Exercise Contexts}
\label{sec:robotCoach}
A rather recent interdisciplinary project in the area of using IVAs in the context of physical exercise is the Intelligent Coaching Space project \cite{kok2017}. Here, a critical aspect is instruction and demonstration by virtual coaches, which clearly falls into the category of fitness coaching via modern technological means. 

There have been two previous attempts to use robots as fitness instructors. One was used in an astronautic setting \cite{2617293,sussenbach2014robot} whereas the other was aimed at fitness aspects in elderly care \cite{Fasola2011,Fasola2013c}. The former was highly focused on motivational aspects of the robot's presence whereas later work, building on the same research, introduced aspects of the robot -- a NAO robot -- joining in on exercises~\cite{schneider2016motivational}. 

Previous work has incorporated a NAO robot into a scenario with cardiac rehabilitation patients \cite{lara2017human}. In this scenario, a NAO robot was used as a tool to help motivate subjects undergoing therapeutic exercises. While this paper showed the robot could be incorporated into a practical health-care task in a sensible manner, it limited the robot's tasks to asking about the Borg scale level (measuring the perceived activity difficulty) and giving feedback based on exercise parameters. Our survey seeks to further explore the role of the robot within the scope of exercise and evaluate how the perception of the robot could potentially improve a subject's performance.

\subsection{Measuring User Preferences}
We aim to investigate whether a robot coach or companion is the preferred role. To this end, we employ three measurement instruments. 

The first questionnaire which we incorporated into our overall questionnaire is the Coach-Athlete Relationship Questionnaire (CART-Q) \cite{jowett2004coach}. It measures the quality of a relationship between an athlete and a coach. 

The second instrument which we deploy is a version of the mind perception questionnaire. This has previously been used in other studies in human-robot interaction (HRI) \cite{gray2007dimensions}. We employ a shorter version of the instrument as used in Gray et al. \cite{gray2011distortions}. Mind perception has previously been used in studies relating to the ``uncanny valley'' effect \cite{gray2012feeling}, motivational states in the context of human-robot interaction (HRI) \cite{eyssel2013}, friendliness in dialogues with intelligent virtual agents \cite{Griffiths2015}, or studies on how adolescence perceive androids \cite{eyssel2017mind}. 

The third instrument in our study is the System Usability Scale (SUS) \cite{brooke1996sus}. It consists of 10 statements (alternating between positive and negative sentiment) that are scored on a 5-point scale of agreement where higher scores represent better usability. 

Using all three instruments allows us to gain some insights into the type of relationship users imagine themselves having with the robot during a workout scenario, while also considering aspects that are overall important for interaction, such as mind perception and system usability.

\section{Methodology}
We constructed an online questionnaire using LimeSurvey. 
The questionnaire is supposed to be stand-alone, and no direct contact with a robot could be provided to the respondents. Therefore, we tried to paint as vividly as possible a picture of our NICO robot \cite{kerzel2017nico} as a partner in exercise contexts. 


\subsection{NICO - the Neuro-Inspired COmpanion}
NICO, the Neuro-Inspired COmpanion \cite{kerzel2017nico}, is a robotic research platform for multimodal human-robot interaction and embodied neuro-cognitive models. NICO features a flexible, modular open source design, that can be adapted to novel research questions and experimental setups \footnote{Visit http://nico.knowledge-technology.info for further information and video material.}. 
NICO stands about 1 meter tall and weighs about 7 kilograms; it falls into the category of child-sized robots, the size was carefully chosen to enable a safe, non-threatening, and approachable design while also enabling the robot to interact with a complex domestic environment \cite{Eppe2017}.

\begin{figure}
  \includegraphics[width=\columnwidth]{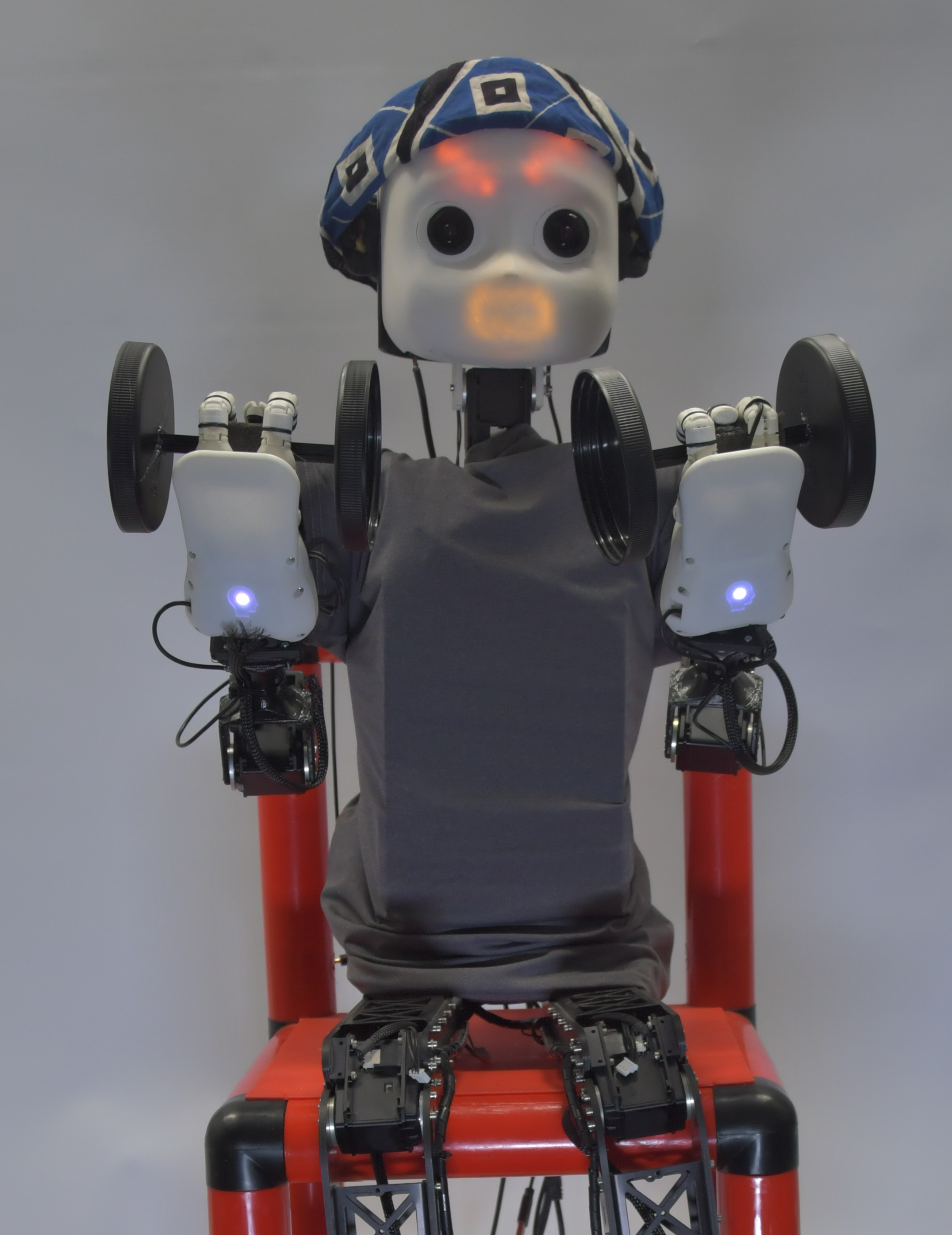}
  \caption{The NICO robot doing biceps curls. The LED facial features of the NICO allows for mimicking of strained or relaxed expressions that are recognizable by humans.}
  \label{fig:teaser}
\end{figure}


The physical design of NICO allows a wide range of motion for 30 degrees of freedom, exceeding the normal limits of human motion, e.g. the elbow can bend over 90 degrees backwards. However, the API developed to control NICO allows to constrain these motions on a software level to human limits to create human-like motion patterns.

NICO's approachable companion role is further underlined by the robot's abilities to both display and recognize facial expressions: Three freely programmable LED arrays are integrated into NICO's head behind the eye and mouth areas, which can be used to display stylized facial expressions \cite{Churamani2017}. In a related study, participants evaluated NICO using the Godspeed questionnaire \cite{bartneck2009measurement}, the ability of NICO to display facial expression increased the subjective ratings of NICO by the users in all five categories of anthropomorphism, animacy, likability, perceived intelligence and perceived safety \cite{kerzel2017nico}. Furthermore, neuro-cognitive models were developed that utilize NICO's embedded visual sensors to learn to recognize general and user-specific emotion expressions \cite{Churamani2017}.

%

\subsection{The Context Provided on the Questionnaire}
On the landing page of the questionnaire, we described our goal of investigating robots deployed in physical exercise settings. On the following page, we introduced our NICO robot via a short text and a video of NICO in interaction with a person. This multimedia content was particularly aimed at facilitating the respondents' imagination with respect to the interactive component.

The video was created in previous work for a study focused on social interaction and personal robotics \cite{NABCFHMMNNSSGHNSTWW17,CABFHMMNNNSSGHNSTWW17}. We extracted a short fifteen-second clip from this video which only shows the robot and user introducing themselves to each other and engaging in a social dialogue. 
After the video, participants were introduced to the exercising scenario. We prepared a number of visual stimuli in the form of pictures of NICO performing weight lifting exercises. Specifically, we showed pictures of biceps curls, one-arm triceps extensions, dumbbell shoulder presses, and seated lateral raises. 
To ensure that the participants actually paid attention to these stimuli and could vividly imagine the robot performing the exercises, we asked questions which the participants had to answer on each of these exercises.  For example, in the context of biceps curls we asked "How many kilos do you think NICO can lift with each arm?" These questions served no actual purpose with respect to the research question, but were useful to ensure that participants actually imagined how NICO would perform in a gym.

Finally, we included the main critical stimulus with respect to the manipulation. This was a paragraph describing NICO's role within the scenario, followed by a picture of the NICO robot performing another exercise. The descriptive text was based on the description of the KISMET robot used by Gray, Gray and Wegner \cite{gray2007dimensions} in their study of mind perception: 

\textit{NICO is part of a new class of ``sociable'' robots that can engage people in natural interaction. To do this, NICO perceives a variety of natural social signals from sound and sight and delivers its own signals back to the human partner through gaze direction, facial expression, body posture, and natural speech.}

\begin{figure*}
  \includegraphics[width=\textwidth]{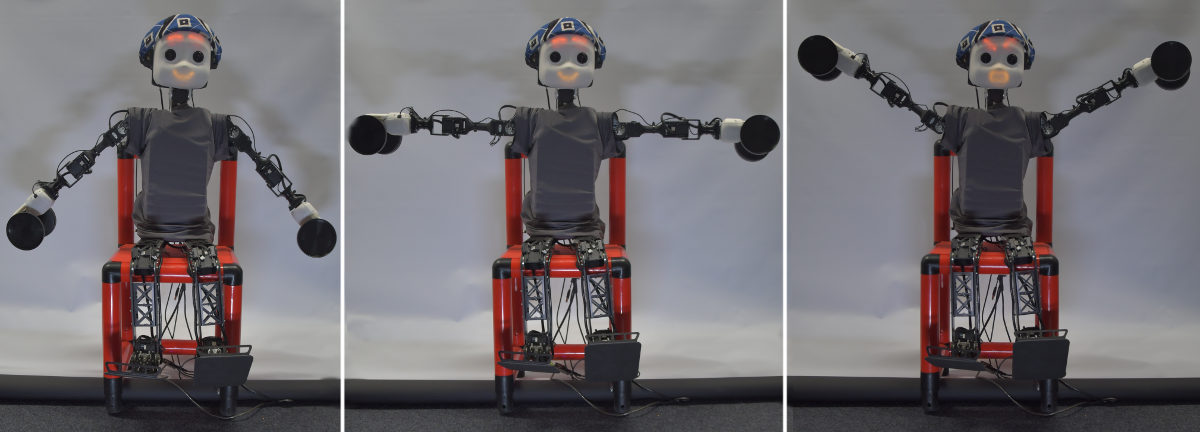}
  \caption{Survey participants were shown static images of the NICO performing individual stages of different exercises in sequence. These stages were placed in sequence in order to give participants the impression of the NICO actually performing the entire exercise. The depicted sequences act as a basis for participants answering questions about imagined aspects of the robot. }
  \label{fig:latraise}
\end{figure*}

The critical manipulation was administered in the form of two possible sentences following this description which distinguished between a fitness companion and a fitness coach. These two possible sentences were:
\begin{itemize}
\item \textit{In the current scenario, NICO will be used as a \textbf{fitness companion}. He reminds users of their intended exercise schedule and does the exercises together with them.}
\item \textit{In the current scenario, NICO will be used as a \textbf{fitness coach}. He reminds users of their intended exercise schedule, shows them exercises to copy and corrects mistakes in their movements.}
\end{itemize}
An example of how the NICO may act in either of these roles can be seen in Figure \ref{fig:latraise}, where the NICO performs an exercise and gives an impression of strain at certain points in the exercise. After the aforementioned statements, the questionnaire presented four sets of questions which will be discussed below.

\begin{figure}
  \includegraphics[width=\columnwidth]{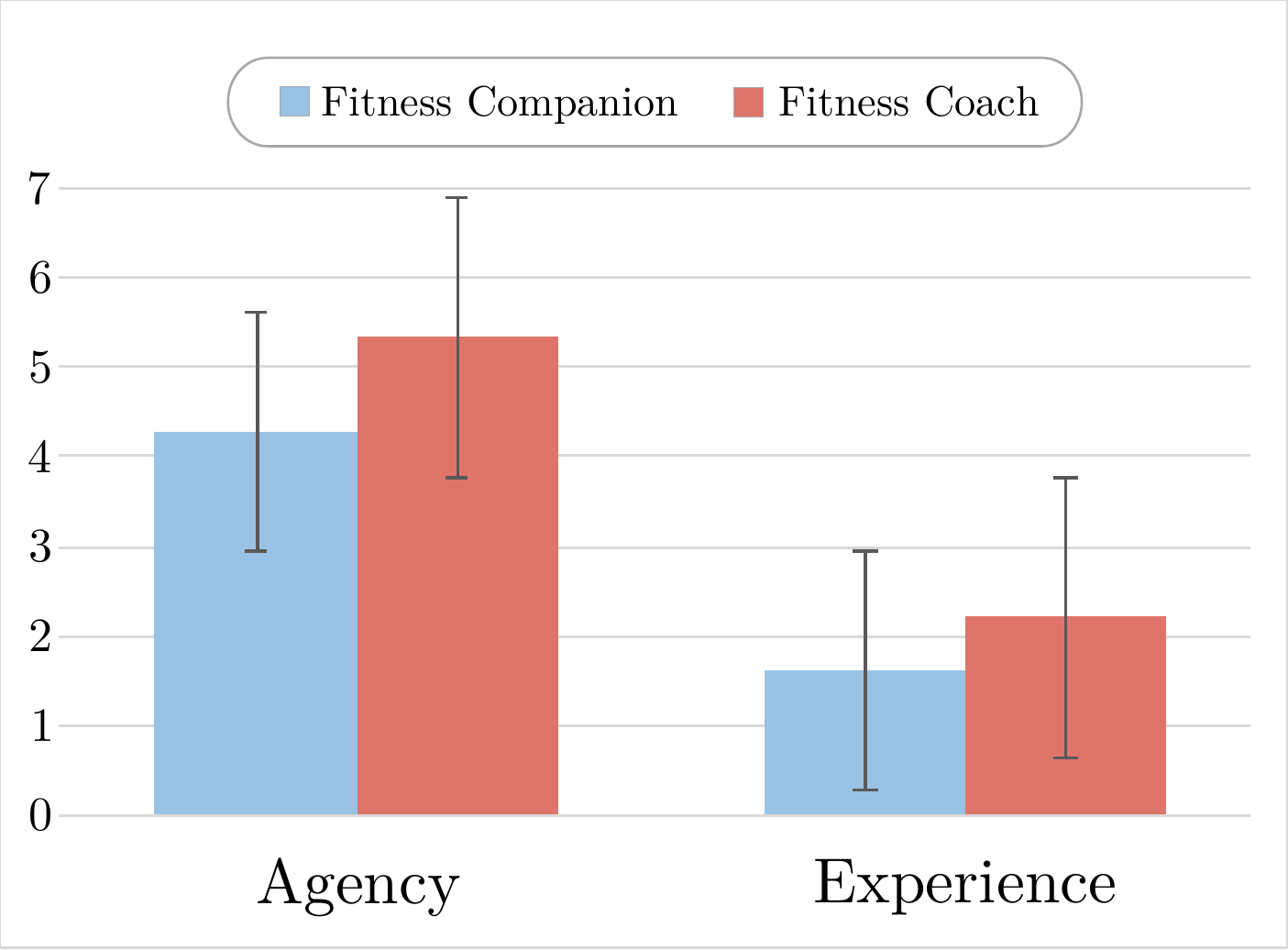}
  \caption{Results for the mind perception questionnaire. Shown are average scores and standard error for the factors Agency and Experience, respectively grouped by the companion (red) and coach (blue) conditions.}
  \label{fig:mind}
\end{figure}

\section{Preliminary Results}
In the following, we present our results with respect to the questionnaire. While more than 70 participants started the survey, only 16 of them completed the questionnaire. Hence, we considered only those 16 participants who completed the survey (\textit{N} = 16). Due to the low number of considered participants, we treat these results as indicative and preliminary. 
While \textit{N} = 16 is to be considered too small for firm statements, a number of HRI studies, such as the one by Baxter et al. \cite{baxter2016characterising}, do indeed work with small sample sizes, which would probably be viewed with skepticism in other disciplines. 
We make no attempt at performing null-hypothesis testing or inferential statistics but show the descriptive statistics as recommended by some current sources \cite{Trafimow2015,baxter2016characterising}. We plan to increase the sample size in future work. The results here are treated as merely indicative.

Our questionnaire was designed to look at three issues, namely (1) the robot as a ``character'' in terms of an entity which can display features of having a mind, (2) the robot as a coach versus the robot as a companion during training, and (3) the robot as a machine. The results will be discussed below. We see user preferences here not as absolute but as an interesting means of seeing which version might be preferred for which context. 

\subsection{Mind Perception}
The overall results for the mind perception questionnaire can be seen in Figure \ref{fig:mind}.
The values for \textbf{agency} in the mind perception questionnaire are slightly higher in the coach condition than in the companion condition. The coach condition has a maximum value of 7 and a minimum value of 1 while the companion condition shows the same maximum (7) or minimum values (1). The mean of the coach condition is \textit{M} = 5.33 (\textit{SD} = 2.00) and the mean of the companion condition is \textit{M} = 4.29 (\textit{SD} = 2.45). The standard deviation suggests that there is more variation in the values for the companion condition. The mean values overall suggest that, within the population on which our system was tested, the coach condition was rated slightly more usable than the companion condition. 

The values for \textbf{experience} in the mind perception questionnaire are slightly higher in the coach condition than in the companion condition. The coach condition has a maximum value of 5 and a minimum value of 1 while the companion condition has a maximum value of 6 and a minimum value of 1. The mean of the coach condition is \textit{M} = 2.22 (\textit{SD} = 1.45) and the mean of the companion condition is \textit{M} = 1.62 (\textit{SD} = 1.57). The mean values overall suggest that, within the population on which our system was tested, the coach condition was rated slightly more usable than the companion condition. 

Overall, the participants attributed more ``mind'', as a composite construct of agency and experience to the coach condition. 

\subsection{CART-Q}
The \textbf{CART-Q} questionnaire was used in order to see whether people would imagine their relationship with NICO as a coach or fitness companion to be effective. The CART-Q is usually used to assess relationships between athletes and coaches. While the questions are formulated differently for the two cases, the content of the questions is the same. Therefore, we can report these two question sets here, in an abbreviated but succinct manner by contrasting the means of the two question sets. 

\begin{figure}
  \includegraphics[width=\columnwidth]{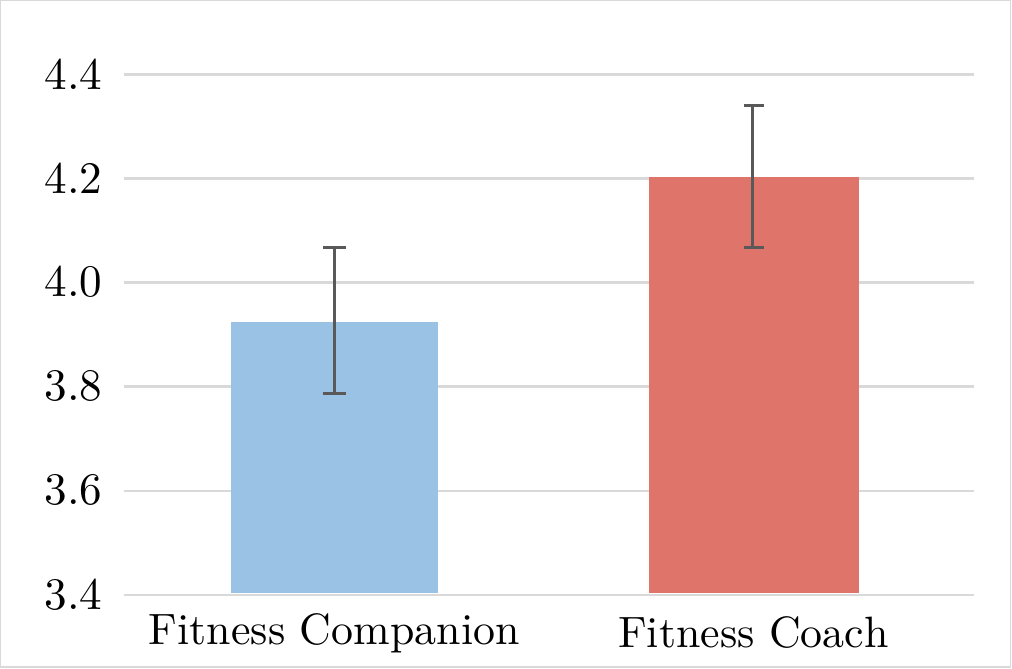}
  \caption{Results for the Coach-Athlete Relationship questionnaire (CART-Q). Shown are average scores and standard error for the companion (red) and coach (blue) conditions.}
  \label{fig:cart}
\end{figure}

The overall results can be seen in Figure \ref{fig:cart}. The values for the CART-Q are slightly higher in the coach condition than in the companion condition. The coach condition has a maximum value of 7 and a minimum value of 1 while the companion condition shows the same maximum (7) or minimum values (1). The mean of the coach condition is \textit{M} = 4.17 (\textit{SD} = 1.50) and the mean of the companion condition is \textit{M} = 3.92 (\textit{SD} = 1.89). The standard deviation suggests that there is more variation in the values for the companion condition. The mean values overall suggest that, within the population on which our system was tested, the coach condition was rated slightly more usable than the companion condition. 

\subsection{System Usability Scale}
\begin{figure}
  \includegraphics[width=\columnwidth]{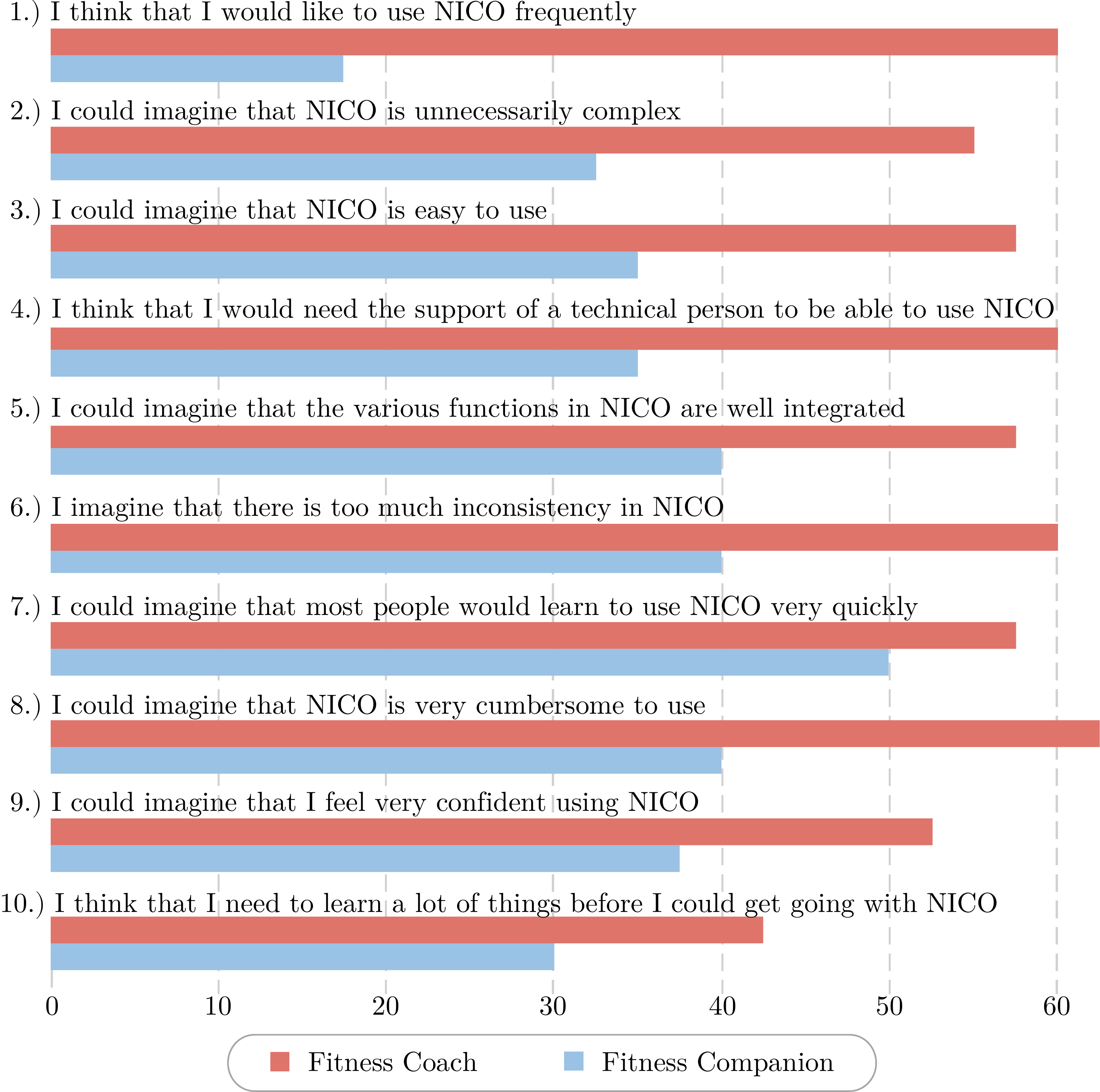}
  \caption{Average SUS scores per statement, grouped by fitness coach (red) and fitness companion (blue). The presented statements are alternating between positive (questions 1, 3, 5, 7, 9) and negative (2, 4, 6, 8, 10) sentiment. A higher score equals a higher usability, regardless of the statement being positive or negative (agreement is inverted for negative statements within the SUS score).}
  \label{fig:sus}
\end{figure}

The average \textbf{System Usability Scale} (SUS) scores per presented statement are shown in Figure \ref{fig:sus}.
The values are higher in the coach condition than in the companion condition. The coach condition has a maximum value of 62.5 and a minimum value of 42.5 while the companion condition has a maximum value of 50 and a minimum value of 17.5. The mean of the coach condition is \textit{M} = 56.50 (\textit{SD} = 5.67) and the mean of the companion condition is \textit{M} = 35.75 (\textit{SD} = 8.42). The standard deviation suggests that there is more variation in the values for the companion condition. The mean values overall suggest that, within the population on which our system was tested, the coach condition was rated more usable than the companion condition. 
\section{Discussion and Future Work}
The results of this pilot-like study show that, within the population tested (\textit{N} = 16), the coach condition was rated slightly higher with respect to all three measures of ``displaying a mind'', being a good sports partner and in terms of being more usable. This is an interesting result that can be used in future work. 

The limitations of the current work are clearly related to the small sample size. The results cannot be generalized to a larger population. However, this was a first test of a composite questionnaire which included a measure from social psychology, an instrument from exercise science and a commonly used indicator of usability. 


In future work, we intend to incorporate both cooperative and adversarial personality traits into a robot and measure the influence this has on users performing exercises in real-world scenarios. The goal of cooperative behaviors exhibited by the robot would be to support and motivate the user while more adversarial traits would aim to inspire and challenge the user. This would follow in the line of work suggested by \citet{lara2017human} of further exploring the possibilities presented by the presence of a robot. By examining how the robot adopting different personas affects user performance and experience of doing various exercises, we hope to gain a better understanding of how robots can have a positive and meaningful role in cooperative tasks.

\section{Conclusion}
We tested the effects of asking people to imagine exercising with a robot. Within the existing work, we identified technology which acts as an instructional tool versus such technologies being deployed as motivational aids. Therefore, the scenario we asked people to imagine was a robot either being a fitness companion or fitness coach. 
In this study, our NICO robot was preferred by the participants when they imagined him being a coach. We tested how much mind people attribute to NICO in these two scenarios, how they see their relationship to the imagined exercise partner and how usable they would think the system could be. In conclusion, the coach system was rated overall higher with respect to all three measures when compared to the fitness companion. 
It must be noted that the imagined contact might actual have been the main drawback. The criticism here can be twofold. First, interacting with an actual system may be completely different from merely looking at pictures of a robot and reading textual descriptions. Thus, this must be seen as the main limitation which we plan to rectify in the future. The abstract nature of the task might have also resulted in the small sample size. Some participants commented on the fact that they found it hard to imagine the exercise situation with the robot. Thus, this may have actually made some potential participants not finish the questionnaire. 

However, these results are merely indicative and cannot be generalised beyond the small population examined. In the future, we plan to use the same methodology in a laboratory study. For this purpose, we will build a dialogue system of the NICO robot that is able to portray the two personas. 


\begin{acks}
  The authors were supported in preparing the questionnaire by Hwei Geok Ng. Di Fu and Nicoletta Xirakia provided valuable feedback on a first draft of the questionnaire. German I. Parisi provided valuable feedback on a draft of the paper. 
  
	This work was partially funded by the German Research Foundation (DFG) in project Crossmodal Learning (TRR-169), the European Union's Horizon 2020 research and innovation programme under the Marie Sklodowska-Curie grant agreement No 721619 (SOCRATES), and the Hamburg Landesforschungsf\"orderungsprojekt CROSS.

\end{acks}

\bibliographystyle{ACM-Reference-Format}
\bibliography{sample-bibliography} 

\end{document}